\documentclass{article}
\usepackage{ijcai19}

\usepackage{times}
\usepackage{soul}
\usepackage{url}
\usepackage[hidelinks]{hyperref}
\usepackage[utf8]{inputenc}
\usepackage[small]{caption}
\usepackage{graphicx}
\usepackage{amsmath}
\usepackage{amsfonts}
\usepackage{booktabs}
\usepackage{algorithm}
\usepackage{algorithmic}
\usepackage{color}
\usepackage{array}
\usepackage{multirow}
\usepackage{subfig}
\usepackage{times}
\usepackage{epsfig}
\usepackage{amssymb}

\newcolumntype{L}[1]{>{\raggedright\let\newline\\\arraybackslash\hspace{0pt}}m{#1}}
\newcolumntype{C}[1]{>{\centering\let\newline\\\arraybackslash\hspace{0pt}}m{#1}}
\newcolumntype{R}[1]{>{\raggedleft\let\newline\\\arraybackslash\hspace{0pt}}m{#1}}
\newcommand{\ie}{i.e.}
\newcommand{\eg}{e.g.}

\title{Talking Face Generation by Conditional Recurrent Adversarial Network}

\author{
	Yang Song$^{1,}$\thanks{equal contribution, $\dagger$ corresponding author.}
	\and
	Jingwen Zhu$^{2,*}$\and
	Dawei Li$^2$\and
	Andy Wang$^{3,\dagger}$\And
	Hairong Qi$^1$
	\affiliations
	$^1$University of Tennessee,Knoxville\\
	$^2$Samsung Research America\\
	$^3$IBM
	\emails
	\{ysong18, hqi\}@utk.edu,
	\{jingwen.z, dawei.l\}@samsung.com,
	visionxiaolong@gmail.com
}

\begin{document}
	
	\maketitle
	
	\begin{abstract}
		Given an arbitrary face image and an arbitrary speech clip, the proposed work attempts to generate the talking face video with accurate lip synchronization. 
		Existing works either do not consider temporal dependency across video frames thus yielding abrupt facial and lip movement or are limited to the generation of talking face video for a specific person thus lacking generalization capacity. 
		We propose a novel conditional recurrent generation network that incorporates both image and audio features in the recurrent unit for  temporal dependency. 
		To achieve both image- and video-realism, a pair of spatial-temporal discriminators are included in the network for better image/video quality. Since accurate lip synchronization is essential to the success of talking face video generation, we also construct a lip-reading discriminator to boost the accuracy of lip synchronization. 
		We also extend the network to model the natural pose and expression of talking face on the Obama Dataset. 
		Extensive experimental results demonstrate the superiority of our framework over the state-of-the-art in terms of visual quality, lip sync accuracy, and smooth transition pertaining to both lip and facial movement.
	\end{abstract}
	
	\section{Introduction}
	The talking face generation problem aims to synthesize naturally looking talking face video provided with a still facial image and a piece of audio speech. Aside from being an interesting topic from a research standpoint, it has
	a wide-range of applications, including, for example, video bandwidth reduction~\cite{suwajanakorn2017synthesizing}, face animation, and other entertainment applications.
	
	Nevertheless, the talking face generation problem faces extreme challenges as can be summarized from the following three perspectives. 
	First, video generation is, in general, more challenging than still image generation since human visual system is more sensitive to temporal discontinuities in videos than spatial discontinuities in images. Second, audio speech to video generation poses extra rigid requirement on the accuracy of lip synchronization. Third, due to the large variations in human pose, talking speed and style, obtaining a video generation model with generalization capacity for unseen audio and face images is quite difficult.  
	
	Most existing works simplify the video generation problem as a temporal-independent image generation problem, e.g.,  \cite{chung2017you,karras2017audio,zhang2017age}, where the temporal dependency in content (\ie, face) is largely left out and the coarticulation effect cannot be adequately modeled.
	%
	Other works do model the temporal dependency for smooth results. For example, \cite{suwajanakorn2017synthesizing} modeled the dynamics of audio features through recurrent neural network (RNN). By measuring the similarity between the given probe audio feature and the gallery audio feature set which are extracted from the source video set, they can find the best matching mouth region. The matched mouth sequence and the target video are then re-timed and synthesized into the output video.
	Although the result seems very promising, it only works for a given person which largely restricts the generalization capability. In addition, this method only models the lip and mouth region without considering the expression or head pose variations as a whole.

	Compared with these approaches, the proposed framework incorporates both image and audio in the recurrent unit to achieve temporal dependency in the generated video on both facial and lip movements, such that smooth transition across different video frames can be realized. We also observe that the image and audio features (or hybrid features) learned by minimizing the reconstruction error between the generated and ground truth frames are insufficient to accurately guide lip movement. This is  because the reconstruction error only calculates the averaged pixel-wise distance, instead of semantically penalizing inaccurate lip movements. In order to guide the network to learn features related to semantic lip movements, a lip-reading discriminator is adopted to train the network in an adversarial manner.
	In addition, we deploy a spatial-temporal discriminator to improve both photo-realism and video-realism. Compared to previous approaches, no extra image deblurring or video stabilization procedure is needed in our framework. Our method can also be extended to model single person video with natural pose and expression. Instead of only using the hybrid feature to feed into the next recurrent unit, we also include the previously generated image frame such that the natural pose and expression of talking face can be intrinsically modeled.
	Our contributions are thus summarized as follows:
	\begin{itemize}
		\item{We propose a novel conditional recurrent generation network that incorporates both image and audio in the recurrent unit for temporal dependency such that smooth transition can be achieved for both lip and facial movements.} 
		\item{We design a pair of spatial-temporal discriminators for both image-realism and video-realism.}
		\item{We construct a lip-reading discriminator to boost the accuracy of lip synchronization.}
		\item{We extend the network to model the natural pose and expression of talking face on the Obama Dataset.}
	\end{itemize}
	
	\section{Related Work}
	\subsection{Speech Face Animation}
	In recent years, several new developments have been reported, \eg, \cite{suwajanakorn2017synthesizing} trained the lip model for only one person, \ie, President Obama. 
	The trained model, however, is difficult to adapt to other people due to the rigid matching scheme used and the large variation in mouth movement. \cite{karras2017audio} learned a mapping between raw audio and 3D meshes by an end-to-end network. 
	Since this method aims to generate 3D mesh animation, it cannot capture the tongue, wrinkles, eyes and head motion which are essential for image level face animation. \cite{taylor2017deep} learned the mapping between audio and phoneme categories, then matched the best predefined lip region model according to the phoneme label. 
	\subsection{Video Prediction and Generation}
	\label{subsec:related_video_generation}
	Video prediction has been widely studied in many works~\cite{mathieu2015deep,oh2015action,srivastava2015unsupervised,liang2017dual}. It aims to predict the future frames conditioned on previous frames. 
	With the success of GAN~\cite{goodfellow2014generative}, a few works have attempted to generate videos through the adversarial training procedure. \cite{vondrick2016generating} decomposed video generation into motion dynamic and background content generation. 
	However, it is restricted to generate only fixed length videos. 
	\cite{tulyakov2017mocogan,villegas2017decomposing} disentangled the video generation problem as content and motion generation. By fixing the content representation and changing the motion latent variables, the video can be generated with motion dynamic under the same content. 
	However, with the unconditional setting, the generated video suffers from low-resolution and fixed short length issues. 
	To model the temporal consistency, Vid2Vid~\cite{vid2vid} proposed a sequential generation scheme where current frame depends on previously generated frame(s).
	
	\subsection{Concurrent Works}
	Concurrent with our paper, there have been two recent closely related works~\cite{chen2018lip,zhou2019talking}. \cite{chen2018lip} focuses on generating the lip region movement. However, keeping the identity across different frames while preserving the video-realism is challenging. Even if the generated lip region images can be blended into the whole face image, the noticeable inconsistency and the existence of those unrealistic static regions will still affect the video-realism. Compared to~\cite{chen2018lip}, \cite{zhou2019talking} designed different network architectures as well as loss functions such that the whole face is modeled. Nonetheless, the video is generated by stacking sequence of independently generated frames, thus the temporal dynamic is not well modeled which has caused strong inconsistency between frames and quite noticeable ``Zoom in and out'' effect. In this case, a post-processing of ``video stabilization'' has to be applied to generate satisfactory results.
	\section{Proposed Method}
	
	\subsection{Problem Formulation}
	\label{sec:problem}
	We hypothesize that the audio to video generation can be formulated as a conditional probability distribution matching problem. In this case, the problem can be solved by minimizing the distance between real video distribution and generated video distribution in an adversarial manner~\cite{goodfellow2014generative}.
	Let $A \equiv \{A_1, A_2, \cdots, A_t,\cdots, A_K\}$ be the audio sequence where each element $A_t$ represents the audio clip in this audio sequence. Let $I \equiv \{I_1, I_2, \cdots, I_t, \cdots, I_K\}$ be the corresponding real talking face sequence. And $I^{*} \in I $ is the identity image which can be any single image chosen from the real sequence of images.  Given the audio sequence $A$ and one single identity image $I^*$ as condition, the talking face generation task aims to generate a sequence of frames $\tilde{I} = \{\tilde{I_1}, \tilde{I_2}, \cdots, \tilde{I_t}, \cdots, \tilde{I_K}\}$, so that the conditional probability distribution of $\tilde{I}$ given $A$ and $I^*$ is close to that of $I$ when given $A$ and $I^*$, i.e., $p(\tilde{I} | A, I^*) \approx p(I |A, I^*)$.%
	
\begin{figure}[ht]
		\centering
		\subfloat[Frame]{\includegraphics[width=.33\columnwidth]{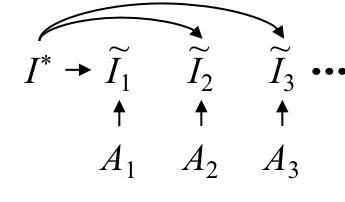}\label{subfig:frame}}
		\subfloat[Sequential]{\includegraphics[width=.33\columnwidth]{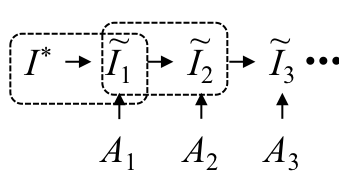}\label{subfig:sequential}}
		\subfloat[Recurrent]{\includegraphics[width=.33\columnwidth]{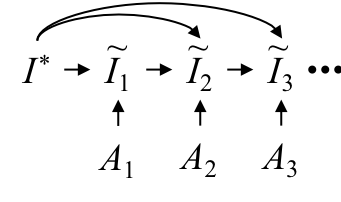}\label{subfig:recurrent}}
		\caption{Illustration of different condition video generation schemes. (a) Frame-to-frame generation scheme (no correlation across different frames). (b) Sequential frame generation scheme (only short term frame correlation is considered). The dash block indicates when $L=2$. (c) Recurrent frame generation scheme where the identity image $I^*$ is fed into all the future frame generation to preserve long-term dependency.}
		\label{fig:schemecmp}
	\end{figure}

	Generally speaking, there have been two schemes developed for conditional video generation, namely, frame-to-frame (Fig.~\ref{subfig:frame}) and sequential frame generation as used in Vid2Vid~\cite{vid2vid} (Fig.~\ref{subfig:sequential}). The frame-to-frame scheme former simplifies the video generation problem into image generation by assuming the i.i.d between different frames. On the other hand, the sequential scheme generates the current frame based on previously generated frame to model the short term dependency. 
	
	For the talking face generation problem in specific where only the audio sequence and one single face image are given, it requires the generated image sequence to 1) preserve the identity across a long time range, 2) have accurate lip shape corresponding to the given audio, and 3) be both photo- and video-realistic. 
	The video generated by the frame-to-frame scheme tends to exhibit jitter effect because no temporal dependency is modeled. The sequential generation scheme cannot preserve the facial identity in long duration because only short term dependency is modeled. 
	To solve these issues, instead, we propose the so-called recurrent frame generation scheme, as shown in Fig.~\ref{subfig:recurrent}, where the  next frame is generated not only depending on the previously generated frames but also the identity frame to preserve long term dependency.

	As illustrated in Fig.~\ref{subfig:recurrent}, another problem with this modeling scheme is that although it considers the image temporal dependency, the audio temporal relation is still ignored. In order to solve these problems, we propose a hybrid recurrent feature generation modeling scheme to capture both the visual and audio dependency over time. We use a recurrent neural network to ingest both the image and audio sequential signal and generate the image sequence directly from a decoder network.  
	We will explain the detailed design in the following section.	
	
	\subsection{Conditional Recurrent Video Generation}
	
	In order to map independent features to a sequence of correlated ones, we apply the recurrent unit on the hybrid features to enforce the temporal coherence on both image and audio signals. 
	Our proposed conditional recurrent video generation network is illustrated in Fig.~\ref{fig:flow}. The inputs to the network are collected by the following feature extraction modules.
	
	\subsubsection{Audio Feature Extraction}
\begin{figure}[t]
		\centering
		\includegraphics[width=1\columnwidth]{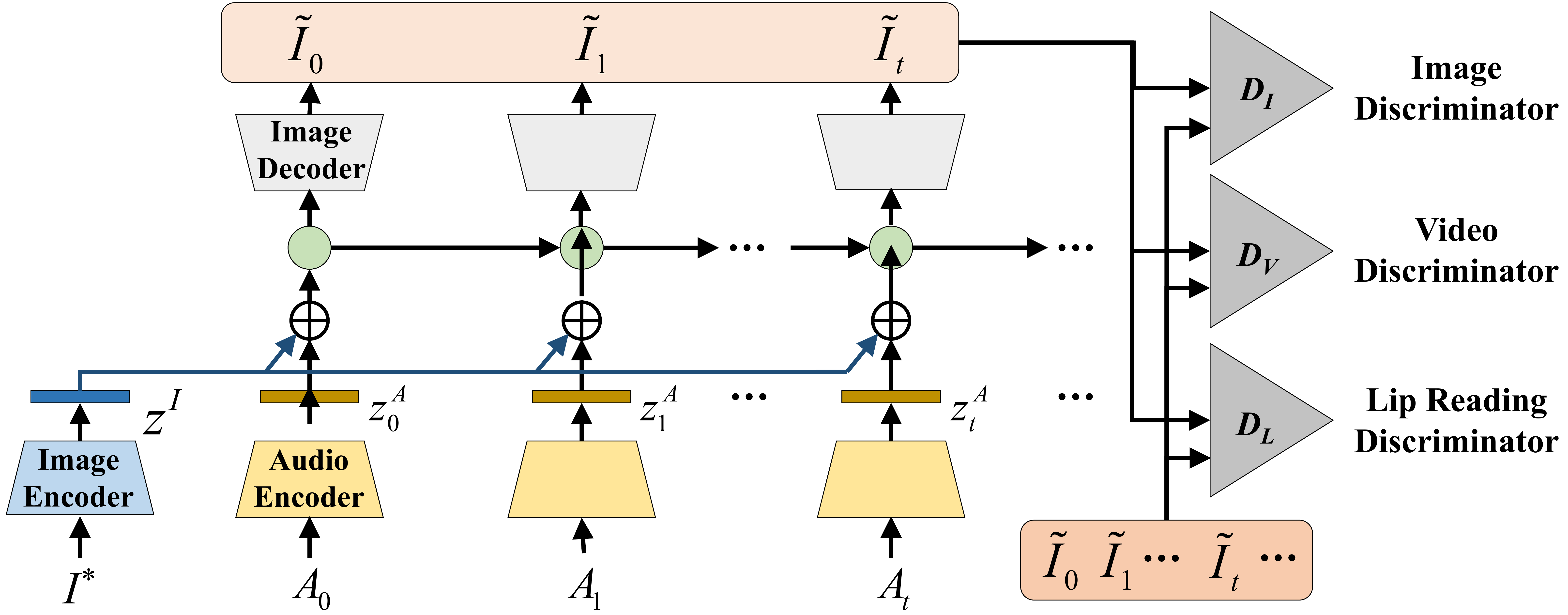}
		\caption{The proposed conditional recurrent adversarial video generation network structure.}
		\label{fig:flow}
	\end{figure}

	Given an entire audio sequence $A$, we use a sliding window to clip the audio file into several audio segments. In Fig.~\ref{fig:flow}, each $A_{t}$ represents the Mel-Frequency Cepstral Coefficients (MFCC) features of each audio segment. We further feed each audio MFCC feature $A_t$ into an audio encoder $E_A$ to extract the audio feature $z^A_t = E_A(A_t)$.\\
	\subsubsection{Image Feature Extraction}
	Frames that have the corresponding lip shapes are extracted according to the starting and ending time of each audio segment $A_t$. There are usually multiple frames correspond to one audio segment, where we simply use the middle one as the image $I_t$ with the corresponding lip shape. The identity image $I^*$ can be randomly selected from $I$. The image feature $z_I$ is calculated by an image encoder as $z_I = E_I(I^*)$.

	A series of audio feature variables denoted as $z^A= [z^A_0, z^A_1, \cdots,z^A_{t}, \cdots, z^A_{K}] $ and the image feature $z^I$ are concatenated to generate a hybrid feature where both the face and audio information are incorporated. 
	The concatenated feature is later fed into the decoder network $Dec$ to generate the target image with desired lip shape while preserving the same identity using the reconstruction loss ${L}_{rec}$.
	
	\begin{equation}
	\mathcal{L}_{rec} = \left\| I_{t} - \tilde{I}_{t} \right\|_1,
	\label{eq:rec_loss}
	\end{equation}
	where $\tilde{I}_{t} =Dec(z^I,z^A)$ = $G(A, I^*)$ and $I_t$ are the generated and ground truth video frames, respectively. 
	\subsection{Adversarial  Learning}	
	\label{subsec:adversarial}
	
	\subsubsection{Lip-Reading Discriminator }
	\label{subsec:sementic_lip_guidance}
	
	The reconstruction error alone is insufficient to accurately guide the lip movement during the training stage because it only calculates the averaged pixel-wise distance, instead of semantically penalizing the inaccurate lip movements. 
	To solve this problem, we use a lip-reading model as a semantic lip guidance, \ie, a lip-reading discriminator, to update the generator in an adversarial manner. 
	The objective function for updating the lip-reading model $D_l$ is shown in Eq.~\ref{eq:loss_D_l}.
	\begin{equation}
	\mathcal{L}_{l} =   \sum y \log \left( D_{l} \left( I_i \right) \right)-\sum y \log \left( D_{l} \left( \tilde{I}_i \right) \right),
	\label{eq:loss_D_l}
	\end{equation}
	where $y$ is the real word label, the output of $D_{l}$ is the predicted probability. $I_i$ and $\tilde{I}_i$ are the real and fake frame sequence where $i=1,2\cdots,K$. By minimizing $\mathcal{L}_l$, the predictions of fake frame sequences through $D_l$ is forced to be misclassified, while the predictions of real frame sequences are pushed toward their true labels.
	
	\subsubsection{Spatial-Temporal Discriminator}
	Both image discriminator and video discriminator are adopted to improve the image quality as well as temporal realism. 
	Although the functionality of the two discriminators overlaps in enhancing image quality, we demonstrate in Sec.~\ref{subsec:ablation} that the image discriminator is more effective in single image enhancement because it  focuses relatively more on each individual frame. Meanwhile, the video discriminator helps achieve smooth transition between frames. 
	
	The image discriminator
	$D_I$ aims to generate realistic images. The corresponding objective is expressed in Eq.~\ref{eq:L_I},
	\begin{equation}
	\begin{split}
	\label{eq:L_I}
	\mathcal{L}_{I} = \; & \mathbb{E}_{I_t \sim p_I}\left[ \log D_I\left(I_t\right) \right] + \\ 
	~ & \mathbb{E}_{\tilde{I}_t \sim p_{\tilde{I}}}\left[ \log \left(1-D_I \left(\tilde{I}_t \right) \right) \right],
	\end{split}
	\end{equation}
	where $p_I$ and $p_{\tilde{I}}$ denotes the distribution of real images in the training videos and the generated images, respectively. 
	
	The video discriminator
	$D_V$ works on a sequence of images/frames to mainly improve the smoothness and continuity between generated image sequences. Eq.~\ref{eq:L_V} shows the objective function.
	\begin{equation}
	\label{eq:L_V}
	\begin{split}
	\mathcal{L}_{V} = \;  & \mathbb{E}_{I_i \sim p_I}\left[ \log D_V\left( I_i \right) \right] + \\
	~ & \mathbb{E}_{\tilde{I}_i \sim p_{\tilde{I}}}\left[ \log \left( 1-D_V \left( \tilde{I}_i \right) \right) \right].
	\end{split}
	\end{equation}
	
	Finally, the total loss for updating our generation network $G$ is
	\begin{equation}
	\label{eq:total_loss}
	\mathcal{L} = \mathcal{L}_{rec} + \lambda_I \mathcal{L}_I + \lambda_V \mathcal{L}_V + \lambda_{l} \mathcal{L}_{l},
	\end{equation}
	where $\lambda_I$, $\lambda_V$, and $\lambda_{l}$ are weighting parameters for the corresponding loss functions, respectively.
	\section{Experimental Results}
	We evaluate the effectiveness of the proposed method on three popular datasets (Sec.~\ref{subsec:datasets}), and demonstrate our advantages over the state-of-the-art works. Specifically, the evaluation is conducted in terms of image quality, lip movement/shape accuracy, and video smoothness. Both qualitative (Sec.~\ref{subsec:qualitative}) and quantitative (Sec.~\ref{subsec:quantitative}) studies demonstrate the superior performance and generality of the proposed method in generating talking face videos\footnote{\scriptsize{Code and More results: \url{https://github.com/susanqq/Talking_Face_Generation}}}. 
	\subsection{Datasets}
	\label{subsec:datasets}
	We use TCD-TIMIT~\cite{harte2015tcd}, LRW~\cite{chung2016lip}, and VoxCeleb~\cite{nagrani2017voxceleb} in our experiments.
	\textbf{TCD-TIMIT} is built for audio-visual speech recognition, where the sentences contain rich phoneme categories, and the videos are captured under well-controlled environment. 
	\textbf{LRW} is collected from the real world accompanied by truth labels (words), and the videos are short, \ie, lasting only a few seconds. 
	\textbf{VoxCeleb} also contains real world videos with large variation in face pose and occlusion/overlap of faces and speech/audio, and longer duration than LRW. For the LRW and VoxCeleb datasets, we filter out those video segments with extreme facial poses or noisy audios in order to generate more stable video result and facilitate the training process. 
	For a fair comparison and  performance evaluation, we split each dataset into training and testing sets following the same experimental setting as previous works~\cite{chung2017you,chen2018lip,zhou2019talking}.

	\subsection{Experimental Setup}	
	\subsubsection{Network Inputs}
	We extract the video frames to make them synchronized with audio segments. For the input/ground truth images, face regions are cropped from the videos and resized to $128\times128$. For the audio inputs, we try different window sizes for MFCC feature and find that 350ms gives the best result.  For the lip-reading discriminator, we only feed the mouth regions in order to avoid other interference, such as facial expressions and large head movements.
	
	\subsubsection{Network Architecture}
	The audio encoder $E_A$, image encoder $E_I$, image discriminator $D_I$, and image decoder $Dec$ are constructed by convolutional or deconvolutional networks. To capture the spatial-temporal information, we use 3D convolution to build the video discriminator $D_V$. In order to preserve the identity, especially for unseen faces during the training, we follow the idea of U-Net to add more low level features to the image decoder $Dec$.  
	
	\subsubsection{Training Scheme}
	We select ADAM~\cite{kingma2014adam} as the optimizer with $\alpha =0.0002$ and $\beta=0.5$ in the experiment. 
	We first train our network without discriminators for 30 epochs, and then add $D_I$, $D_V$, $D_l$ to finetune the network for another 15 epochs. $D_l$ is initialized from a pre-trained lip-reading model. 
	The weights for $\mathcal{L}_I$, $\mathcal{L}_V$, and $\mathcal{L}_l$ are 1e-3, 1e-2, and 1e-3, respectively. 
	Since we don't have word/sentence labels for VoxCeleb and TCD-TIMIT dataset, we only apply $\mathcal{L}_l$ on LRW samples during the training.
	
	\subsection{Qualitative Evaluation}
	\label{subsec:qualitative}
	\subsubsection{Comparison with Other Methods}
	
	\begin{figure}[t]
		\centering
		\rotatebox{90}{
			\begin{tabular}{C{30pt}C{40pt}C{40pt}C{30pt}}
				\scriptsize Ours & \scriptsize \cite{zhou2019talking} & \scriptsize \cite{chung2017you} & \scriptsize \cite{chen2018lip} 
			\end{tabular}
		}
		\includegraphics[width=.9\columnwidth]{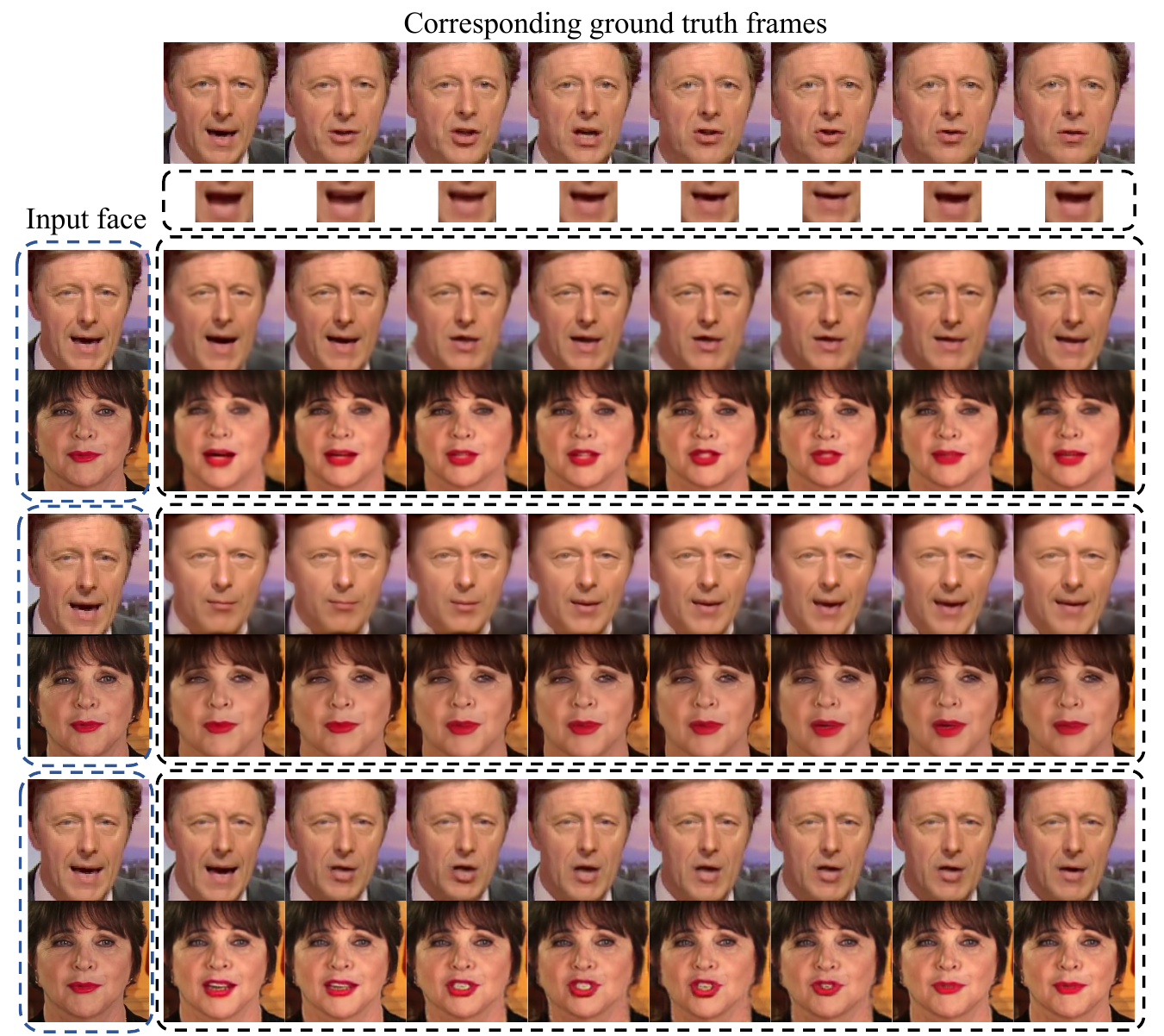}
		\caption{Comparison between the proposed method and the state-of-the-art algorithms. The first row shows the ground truth video, saying ``high edu(cation)'' which is the input audio, and the first image column gives the input faces. \protect\cite{chen2018lip} can only generate the lip region. Frames corresponding to the letters inside parentheses are not presented here.}
		\label{fig:qualtitative_cmp}
	\end{figure}
	
	We qualitatively compare the proposed method with three related works, \ie, \cite{chen2018lip},  \cite{chung2017you} and \cite{zhou2019talking} in this section, as shown in Fig.~\ref{fig:qualtitative_cmp}. 
	The input audio is saying ``high education'', which is randomly selected from the LRW testing set. The corresponding ground truth frames are listed in the first row, and the input faces are shown in the first image column. We feed the same input audio and faces to different methods for fair comparison. Note that \cite{chen2018lip} only works on the mouth region instead of the whole face.
	
	As visualized in Fig.~\ref{fig:qualtitative_cmp}, the frames generated by the proposed method present higher visual quality, \ie, sharper skin texture, more realistic winkles and clearer details (e.g., teeth) than all other methods (please zoom in for better visualization), which demonstrates the effectiveness of using spatial-temporal discriminator as in our framework. The second observation is that we outperform other methods by generating sharper and more discriminative mouth shapes. One reason is due to the application of the lip-reading discriminator. Because of the usage of the lip-reading discriminator, the generator is forced to generate more accurate mouth shapes in order to predict the correct word label. Without such semantic guidance on the lip movement, the generated mouth shapes usually present less expressive and discriminative semantic cues, such as the comparison results listed in Fig.~\ref{fig:qualtitative_cmp} from ~\cite{chung2017you,zhou2019talking}.
	
	In addition, video frames generated by \cite{chung2017you} and \cite{zhou2019talking} contain inter-frame discontinuities and motion inconsistency which are more obvious in the video (provided in the supplement) than in Fig.~\ref{fig:qualtitative_cmp}. The proposed method reduces these discontinuities by performing the recurrent generation to model the temporal dynamics, which has not been considered in other methods.
	

	\subsubsection{Ablation Study}
	\label{subsec:ablation}
	To compare different generation schemes and analyze the effectiveness of different loss functions in our method, we carry out extensive ablation studies as follows.

\begin{figure}[t]
		\centering
		\rotatebox{90}{
			\begin{tabular}{C{27pt}C{23pt}C{22pt}C{22pt}C{20pt}}
				\tiny $\mathcal{L}_{r,I,V,l}$ & \tiny $\mathcal{L}_{r,I,V}$ & \tiny $\mathcal{L}_{r,I}$ & \tiny $\mathcal{L}_{r}$
			\end{tabular}
		}
		\includegraphics[width=0.7\columnwidth]{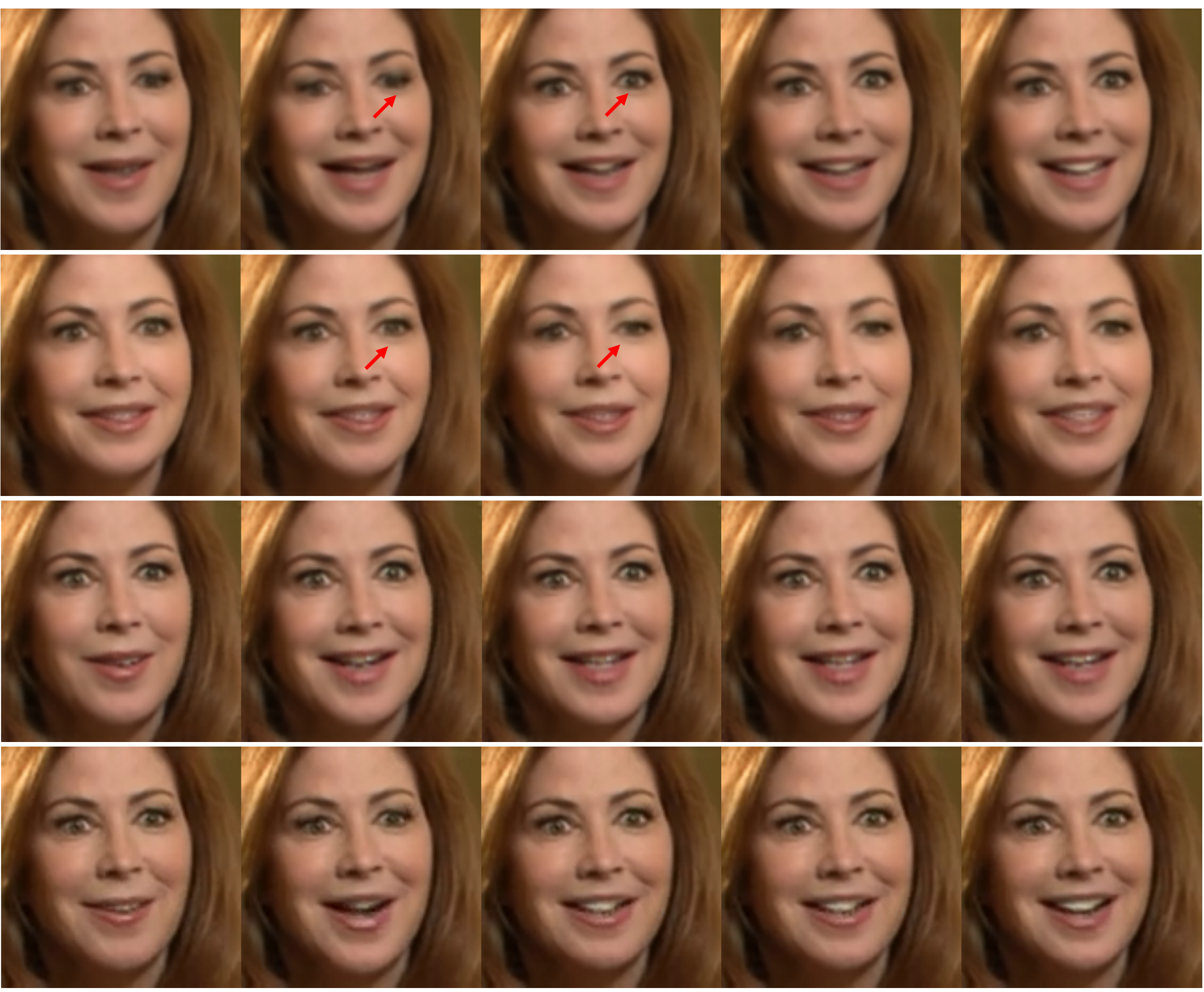}
		\caption{Ablation study on the loss functions used in the proposed method. The rows show the continuous frames generated from the same audio input but different loss combinations as denoted on the left. The subscripts $r$, $I$, $V$, and $l$ indicates $\mathcal{L}_{rec}$, $\mathcal{L}_{I}$, $\mathcal{L}_{V}$, and $\mathcal{L}_{l}$ in Eq.~\ref{eq:total_loss}, respectively. }
		\label{fig:ablation}
	\end{figure}

	Effectiveness of different losses
	in the proposed method (Eq.~\ref{eq:total_loss}) is compared and demonstrated in Fig.~\ref{fig:ablation}, where $\mathcal{L}_r$, $\mathcal{L}_I$, $\mathcal{L}_V$, $\mathcal{L}_l$ represent reconstruction loss, image adversarial loss, video adversarial loss, and lip-reading adversarial loss, respectively. We use the results (first row) produced by only using $\mathcal{L}_r$ as the baseline. From Fig.~\ref{fig:ablation}, we can see that after adding the image adversarial loss $\mathcal{L}_I$, the generated images present more details, \eg, the teeth region becomes much clearer. Adding the video adversarial loss $\mathcal{L}_{r,I,V}$ further sharpens the images (see the teeth region as well) and smooth out some jittery artifacts between frames. Finally, the addition of lip-reading discriminator helps achieve more obvious lip movement. 
	Although the recurrent network in our framework also aims at maintaining the temporal consistency, it plays a more important role on the global video smoothness such as avoiding pose inconsistency (\eg, zoom-in and zoom-out effect). 
	
\begin{figure}[t]
		\centering
		\includegraphics[width=.5\columnwidth]{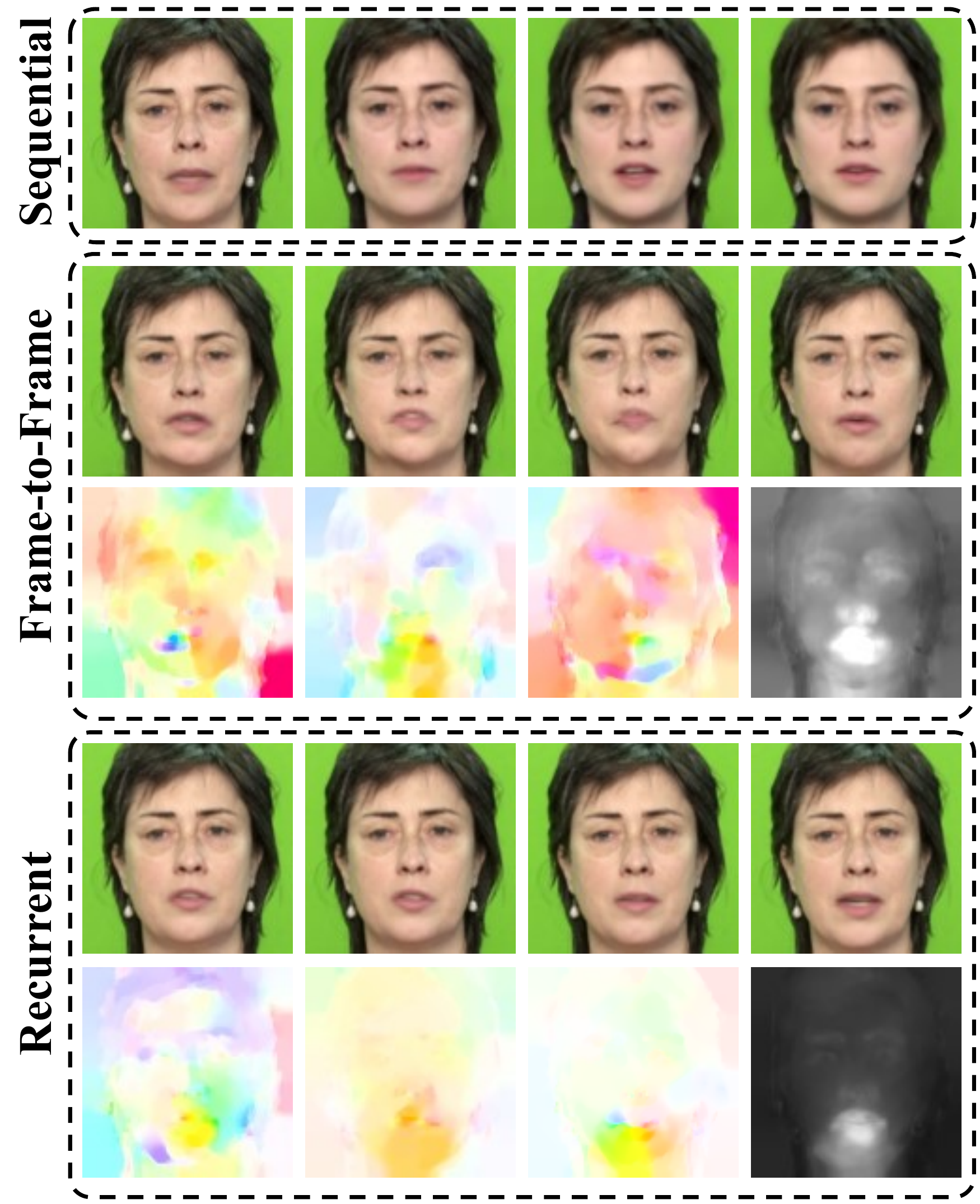}
		\caption{Effect of different generation schemes. The sample is randomly selected from the TCD-TIMIT dataset. From top to bottom:  sequential generation, frame-to-frame generation, and our recurrent generation schemes. Optical flow is calculated on the frame-to-frame and recurrent schemes as shown under the face images. 
		}
		\label{fig:rnnvscnn}
	\end{figure}

	Effectiveness of different schemes is compared in Fig.~\ref{fig:rnnvscnn}. We compare the recurrent generation (bottom block) with other two existing schemes, \ie, sequential generation (top block) and frame-to-frame generation (middle block) as introduced in Sec.~\ref{sec:problem}. Obviously, the sequential generation scheme fails to preserve the identity while the frame-to-frame scheme exhibits large variance between adjacent frames as illustrated by the optical flow. 
	The gray-scale map represents the motion intensity map\footnote{\scriptsize{Assume $(u, v)= optical flow(I_1, I_2)$ is the optical flow between two continuous frames, the motion intensity for each pixel is calculated by $u^2+ v^2$.}} which is calculated by averaging the optical flow for the whole sequence, where brighter pixels illustrate larger variation between adjacent frames.
	Compared with frame-to-frame generator, the recurrent scheme preserves the identity information well and achieves the smooth flow between frames, \ie, most movements are around the mouth area.  
	
	\subsection{Quantitative Evaluation}
	\label{subsec:quantitative}
	
	The same as current related works \cite{chung2017you,zhou2019talking}, we use PSNR and SSIM to evaluate the image quality. To measure the lip movement accuracy, we also use lip-reading accuracy~\cite{chung2016lip} and Landmark Distance Error (LMD)~\cite{chen2018lip} as the measuring criteria to understand the lip movement from the semantic level and pixel level respectively.  Note that \cite{chen2018lip} is excluded in our quantitative study because it only focuses on the lip region which would not be a fair comparison. 
	
	We list quantitative evaluation results as well as the comparison with other methods on the LRW dataset in Table~\ref{tab:cmp}. Our recurrent video generation framework has demonstrated better performance over other frame-by-frame generators.
	
	\begin{table}[ht]
		\centering
		\scriptsize
		\begin{tabular}{l|cccc}
			\hline
			\centering{} & PSNR & SSIM & LMD & LRA \\
			
			\hline
			\cite{chung2017you} & 27.16 &  0.917  & 3.15& $15.2\%$/$28.8\%$   \\  
			\cite{zhou2019talking} & 26.43&  0.894   &5.21   &  $18.2\%$/$38.6\%$ \\
			\hline
			$\mathcal{L}_{r,l,I, V}$&27.43 &0.918  &3.14 & $36.2\%$/$63.0\%$ \\
			$\mathcal{L}_{r,I,V}$&27.43&0.913  &3.14&$18.5\%$/$38.0\%$ \\
			$\mathcal{L}_{r,V}$&27.41 &0.919 &3.10&$18.7\%$/$34.3\%$ \\
			$\mathcal{L}_{r}$&27.77&0.924&3.01 &$18.2\%$/$38.6\%$\\		
			\hline
		\end{tabular}
		\caption{The quantitative evaluation on the LRW testing dataset. The second block lists the results of our recurrent network using different loss functions. LRA (Top1/Top5) denotes the lip-reading accuracy.}
		\label{tab:cmp}
	\end{table}
	
	We also list the result of our method using different loss functions. Our key observation is that by applying discriminators, especially the image and video discriminators, both the PSNR and SSIM values are negatively affected, though they greatly improve the visual quality by adding more details which may not appear in the original images. The reason for this phenomenon is that both PSNR and SSIM  calculate pixel level differences which cannot well reflect the visual quality. This phenomenon has been studied in many super-resolution related works. Similarly, we observe that the best LMD value is achieved by using only the reconstruction loss. We argue that LMD cannot precisely reflect the lip movement accuracy mainly from two aspects. First, for the same word, different people may pronounce it in different ways, \eg, very exaggerated mouth opening range. The generated samples are not necessarily to be the same way as the ground truth. Second, this may be caused by the landmark detection error especially for non-frontal faces. In other words, PSNR, SSIM and LMD cannot accurately reflect whether the generated lip movement is correct or not. 
	
	For better evaluation of the lip sync result, we use the-state-of-the-art deep lip-reading model which is trained on real speech videos to quantify the accuracy. Computer vision based lip-reading models have been used to aid correction of lip movements to improve human's pronunciation~\cite{lip-correction}, and thus should be an ideal metric to justify the authenticity of the generated speech videos. We find with lip-reading discriminator, the lip-reading accuracy is improved from $18.5\%$ and $38.0\%$ to $36.2\%$ and $63.0\%$ on top1 and top5, which approaches the accuracy on real videos ($60\%$ and $80\%$ top1 and top5).
	\subsection{User Study}	
	\label{subsec:userstudy}
	We further conduct the user study for more subjective evaluation.
	We choose Amazon Mechanical Turk (AMT) as the user study platform and all the workers are required to be from either ``UK" or ``USA". 
	In the study, workers are asked to perform pair-wise comparison between videos generated by \cite{chung2017you}, \cite{zhou2019talking}, and ours. 
	Human evaluation is very subjective and different people may focus on different perspectives. 
	Therefore we give clear user instructions, and ask people to evaluate the result from three perspectives: whether the generated video has good smoothness like a real video (realistic of video); whether the lip shape is accurate and  synchronized well with the audio (lip movement accuracy); and whether the image frames are blurred or with artifacts (image quality).	In total, each worker is asked to evaluate 60 test samples from the VoxCeleb dataset and the LRW dataset, each of which has been used to generate 3 videos by three compared methods. 
	For each test sample, videos generated by our method and a compared method are provided to workers, and they are asked to select the better video according to the instruction. Each sample is evaluated by 10 different workers. We summarize the result and list it in Table~\ref{tb:accuracy}.

	\begin{table}
		\centering
		\begin{tabular}{ll}
			\scriptsize
			\begin{tabular}{C{2.4cm}|C{1.34cm}C{1.45cm}C{1.41cm}}
				\hline
				Voting to ours vs. others  & Lip Accuracy&Video Realism & Image Quality \\ \hline
				\cite{chung2017you}   &  $0.74/0.26$   &  $0.66/0.34$ &  $0.73/0.27$  \\ 
				\cite{zhou2019talking}&    $0.68/0.32$ &    $0.87/0.13$   &   $0.70/0.30$    \\ \hline
			\end{tabular}
		\end{tabular}
		\caption{User study of generated videos from proposed method vs other state-of-the-art methods. }
		\label{tb:accuracy}
	\end{table}
	
As shown in Table~\ref{tb:accuracy}, our method clearly outperforms recent approaches~\cite{chung2017you,zhou2019talking}. 
	In specific, our method outperforms the other two methods on lip movement accuracy which is also consistent with the lip-reading model result in Table~\ref{tab:cmp}.
	In addition, our framework is superior on the video realistic metric as well as image quality since other methods suffer from lots of motion artifacts like pose discontinuity and unstable face sizes between frames.

	\section{Study on Natural Pose and Expression}	
\begin{figure}[ht]
		\centering
		\includegraphics[width=.6\columnwidth]{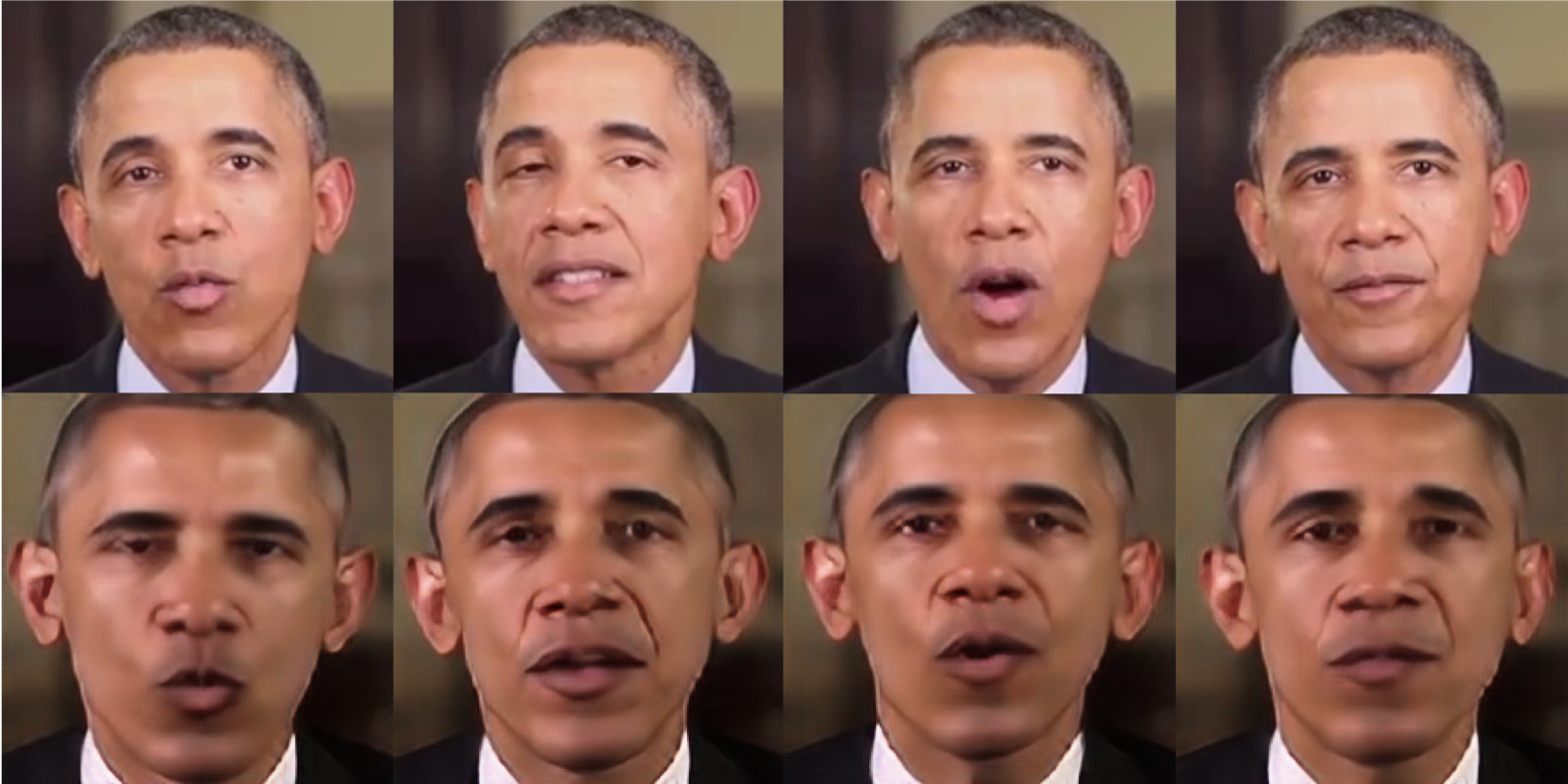}
		\caption{The first row is the ground truth image, the second row is our generated results.}
		\label{fig:obama}
	\end{figure}

	One potential problem for the proposed structure and other concurrent works is that it is difficult to generate the talking face video with natural poses and expressions using 2D method. We observe that instead of only using the hybrid features as inputs to the next recurrent unit, we also include the previously generated image frame such that the natural pose and expression of talking face can be intrinsically modeled. This modification actually integrates the sequential generation scheme and our recurrent generation scheme. Recall that the sequential generation scheme alone will cause changing face issue when generalized to different people, but will not be a problem for single-person video generation.  We train this modified neural network on the Obama dataset~\cite{suwajanakorn2017synthesizing}. As shown in Fig.~\ref{fig:obama}, our generated video exhibits accurate lip shape and natural pose change. 

	
	\section{Conclusion, Limitations and Future Work}
	We presented a new conditional adversarial network for talking face video generation. Our framework mainly utilizes the recurrent adversarial network to capture the temporal dependency of sequential audio features and image information simultaneously. Furthermore, the framework incorporates  three  discriminators to further improve the image quality, video realism and lip movement accuracy in an adversarial training manner. Our extensive experimental results obtained from public constrained and unconstrained data demonstrated the superiority over state-of-the-arts under different performance metrics.  
	
	One of the areas that deserve further investigation is the design of a true end-to-end generation framework, where the raw audio file instead of the MFCC features is used as input to the network. The other area is the incorporation of a sentence-level lip-reading discriminator for guiding better lip motion generation. To improve image quality, the super-resolution~\cite{zhang2019image} work can be incorporated.

	\newpage
	\bibliographystyle{named}
	\bibliography{ijcai19}
	
\end{document}